%% file: icann18.tex
\documentclass[runningheads]{llncs}
\pdfoutput=1
\usepackage{graphicx}
\usepackage{algorithm2e}
\usepackage{multirow}
\usepackage{subfig}
\usepackage{slashbox}
\usepackage{hyperref}
\usepackage{pifont}
\usepackage{color}
\usepackage{hyperref}
\usepackage[capitalize]{cleveref}
\usepackage{enumitem}
\usepackage{array}
\usepackage[font=small,skip=0pt]{caption}
\usepackage{graphicx}
\usepackage[export]{adjustbox}
\SetKw{KwBy}{step} 
\hypersetup{
  pdftitle={Catastrophic forgetting: still a problem for DNNs},    
  pdfauthor={B. Pfülb, A. Gepperth, S. Abdullah, A. Kilian},     
  pdfsubject={Catastrophic forgetting: still a problem for DNNs},   
  pdfcreator={pdflatex},   
  pdfproducer={TeXStudio}, 
  pdfkeywords={DNN, catastrophic forgetting, incremental learning},
  pagebackref=true,
  pdffitwindow=true,
  breaklinks=true,
  letterpaper=true,
  citecolor=red,
  urlcolor=blue,
  colorlinks,
  bookmarks=false
}
\crefname{section}{Sect.}{Sects.}
\crefname{subsection}{Sect.}{Sects.}
\crefname{table}{Tab.}{Tabs.}
\crefname{algorithm}{Alg.}{Algs.}
\renewcommand\tablename{\textbf{Tab.\ {}}}
\renewcommand\figurename{\textbf{Fig.\ {}}}
\renewcommand{\thetable}{\arabic{table}}
\newcolumntype{L}[1]{>{\raggedright\let\newline\\\arraybackslash\hspace{0pt}}m{#1}}
\newcolumntype{C}[1]{>{\centering\let\newline\\\arraybackslash\hspace{0pt}}m{#1}}
\newcolumntype{R}[1]{>{\raggedleft\let\newline\\\arraybackslash\hspace{0pt}}m{#1}}
\begin{document}
\setlength{\parindent}{0cm}
\title{Catastrophic forgetting: still a problem for DNNs}
\author{
B.\ Pf{\"u}lb \and
A.\ Gepperth \and
S.\ Abdullah  \and
A.\ Kilian
}
\authorrunning{B. Pf{\"u}lb, A. Gepperth et al.}
\institute{Fulda University of Applied Sciences, Leipzigerstr. 123, 36037 Fulda, Germany
\email{\{benedikt.pfuelb, alexander.gepperth, saad.abdullah, andre.kilian\}@cs.hs-fulda.de}\\
\url{https://www.hs-fulda.de}}
\maketitle
\begin{abstract}
We investigate the performance of DNNs when trained on class-incremental visual problems consisting of initial training, followed by retraining with added visual classes.
Catastrophic forgetting (CF) behavior is measured using a new evaluation procedure that aims at an application-oriented view of incremental learning.
In particular, it imposes that model selection must be performed on the initial dataset alone, as well as demanding that retraining control be performed only using the retraining dataset, as initial dataset is usually too large to be kept.
Experiments are conducted on class-incremental problems derived from MNIST, using a variety of different DNN models, some of them recently proposed to avoid catastrophic forgetting.
When comparing our new evaluation procedure to previous approaches for assessing CF, we find their findings are completely negated, and that none of the tested methods can avoid CF in all experiments.
This stresses the importance of a realistic empirical measurement procedure for catastrophic forgetting, and the need for further research in incremental learning for DNNs. 
\end{abstract}
\section{Introduction}\label{sec:intro}
\begin{figure}[t!]
  \centering
  \def\svgwidth{0.6\linewidth}
  \subfloat[Training scheme\label{fig:scheme}]{
    \sffamily\large\resizebox{0.33 \textwidth}{!}{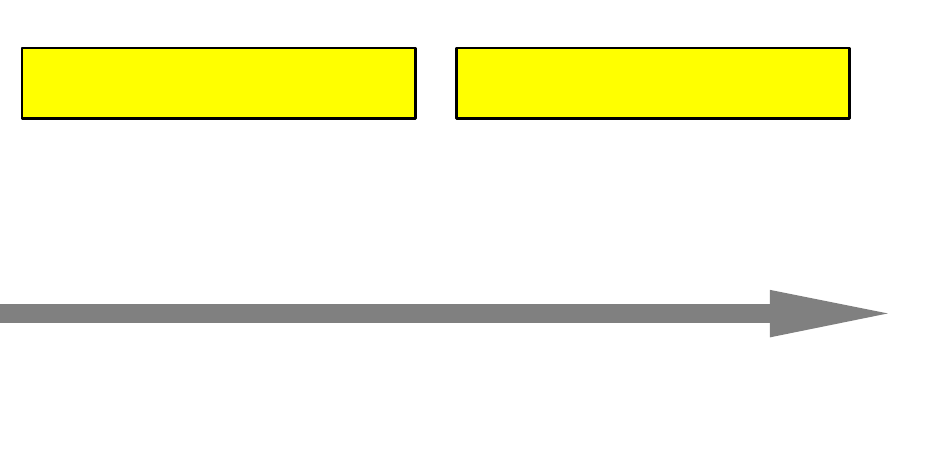} 
  }%
  \subfloat[without CF\label{fig:without_CF}]{\includegraphics[width=0.32\textwidth]{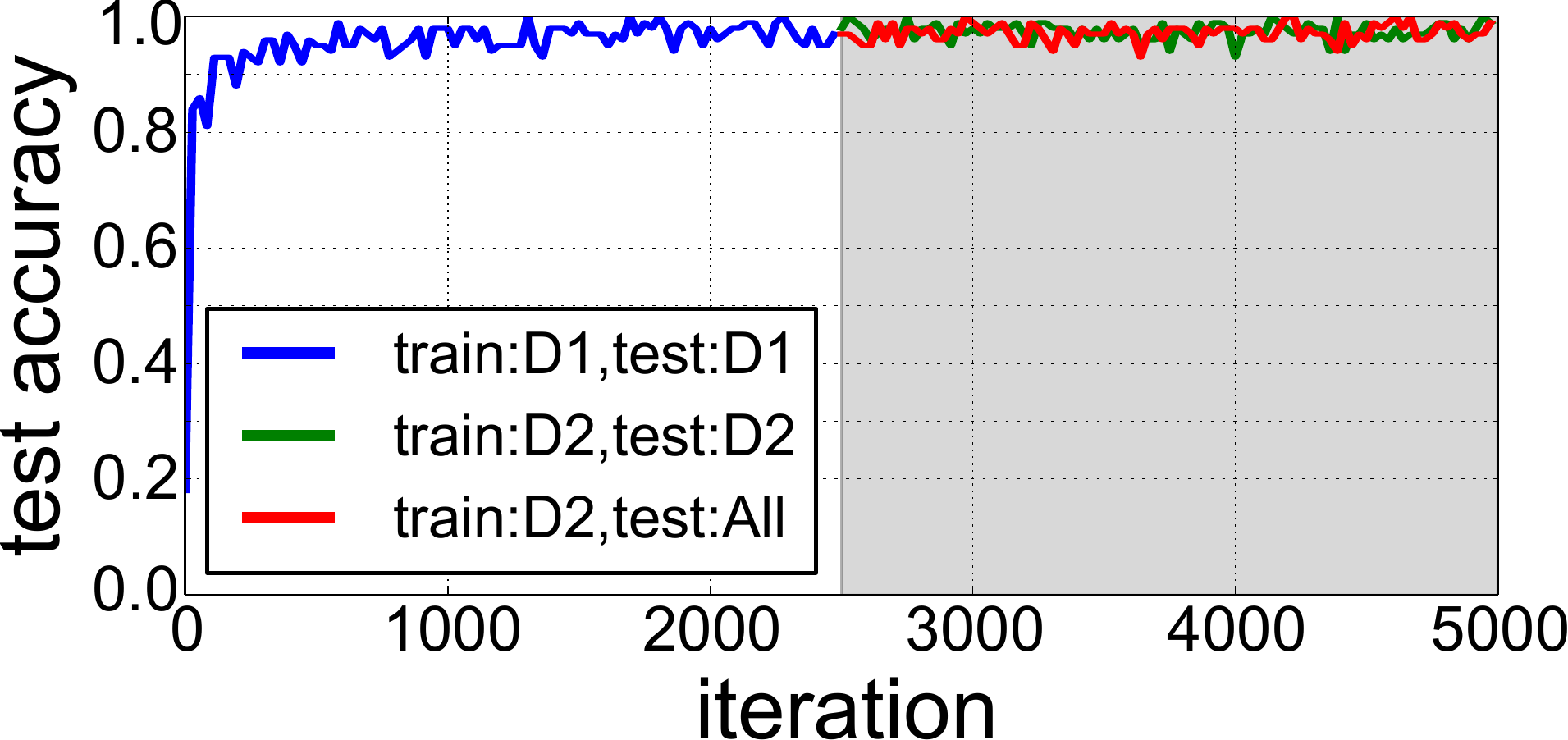}}
  \subfloat[with CF\label{fig:with_CF}]{\includegraphics[width=0.32\textwidth]{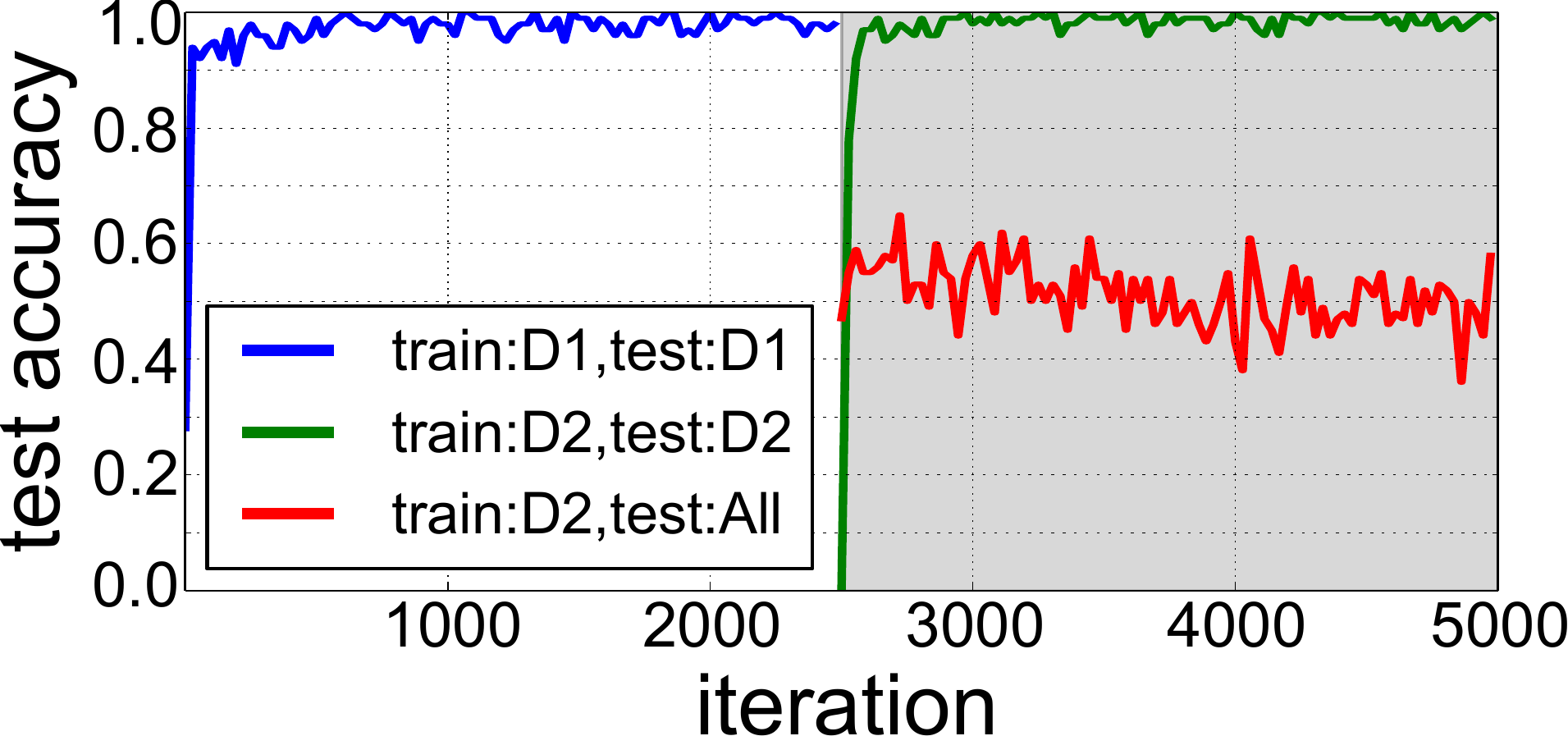}}
  \vspace{0.2cm}
  \normalfont
  \caption[]{
    Scheme of incremental training experiments (see Fig.\ \ref{fig:scheme}) and representative outcomes without and with CF (see Figs.\ \ref{fig:without_CF} and \ref{fig:with_CF}). Initial training with sub-task $D_1$ for $t_{max}$ iterations is followed by retraining on sub-task $D_2$ for another $t_{max}$ iterations.
    During training (white background) and retraining (grey background), test performance is measured on $D_1$ (blue curves), $D_2$ (green curves) and $D_1\cup D_2$ (red curves). The red curves allow to determine the presence of CF by simple visual inspection: if there is significant degradation w.r.t.\ the blue curves, then CF has occurred.
  \label{fig:experiments}
  }
\end{figure}

The context of this article is the susceptibility of DNN to an effect usually termed "catastrophic forgetting" or "catastrophic interference" \cite{french}.
When training a DNN incrementally, that is, first training it on a sub-task $D_1$ and subsequently retraining on another sub-task $D_2$ whose statistics differ (see \cref{fig:experiments}), CF implies an abrupt and virtually complete loss of knowledge about $D_1$ during retraining.
In various forms, knowledge of this effect dates back to very early works on neural networks \cite{french}, of which modern DNNs are a special case.
Nevertheless, known solutions seem difficult to apply to modern DNNs trained in a purely gradient-based fashion.
Recently, several approaches have been published with the explicit goal of resolving the CF issue for DNNs in incremental learning tasks, illustrated in \cite{goodfellow2013empirical,kirkpatrick2017overcoming,srivastava2013compete}.
On the other hand, there are "shallow" machine learning methods explicitly constructed to avoid CF (reviewed in, e.g., \cite{sigaud2011online}), although this ability seems to be achieved at the cost of significantly reduced learning capacity.
In this article, we test the recently proposed solutions for DNNs using a variety of class-incremental visual problems constructed from the well-known MNIST benchmark \cite{lecun-01a}.
In particular, we propose a new experimental protocol to measure CF which avoids commonly made \cite{goodfellow2013empirical,kirkpatrick2017overcoming,srivastava2013compete,lee2017overcoming} implicit assumptions that are incompatible with incremental learning in applied scenarios.
\subsection{Application relevance of catastrophic forgetting}
When DNNs are trained on a single (sub-)task $D_1$ only, catastrophic forgetting is not an issue.
When retraining is necessary with a new sub-task $D_2$, one often recurs to retraining the DNN with all samples from $D_1$ and $D_2$ together. This heuristic works in many situations, especially when the cardinality of $D_1$ is moderate. When $D_1$ becomes very large, however, or many slight additions $D_{(1+n)}$ are required, this strategy becomes unfeasible, and an incremental training scheme (see Fig.~\ref{fig:scheme}) must be used.
Thus, the issue of catastrophic forgetting becomes critically important, which is why we wish to assess, once and for all, where DNNs stand with respect to CF.
\subsection{Approach of the article}\label{sec:intro:approach}
In all experiments, we consider class-incremental learning scenarios divided into two training steps on disjunct sub-tasks $D_1$ and $D_2$, as outlined in \cref{sec:intro} and visualized in \cref{fig:experiments}.
Both training steps are conducted for a fixed number of iterations, with the understanding that in practice retraining would have to be stopped at some point by an appropriate criterion before forgetting of $D_1$ is complete.
The occurrence of forgetting is quantified using classification performance on all test samples from $D_1 \cup D_2$ at the time retraining is stopped (see \cref{fig:experiments} for a visual impression).
In contrast to previous works, our experiments take into account how (class-)incremental learning works in practice:
\begin{itemize}[leftmargin=*,topsep=0.0cm,itemsep=0.1cm]
  \item $D_2$ is not available at initial training
  \item $D_1$ is not available at retraining time as it might be very large.
\end{itemize}
\vspace{1em}
This training paradigm (which we term "realistic") has profound consequences, most importantly that initial model selection has to be performed using $D_1$ alone, which is in contrast to previous works on CF in DNNs \cite{goodfellow2013empirical,kirkpatrick2017overcoming,srivastava2013compete}, where $D_1\cup D_2$ is used for model selection purposes.
Another consequence is that the decision on when to stop retraining has to be taken based on $D_2$ alone.
\par
In order to reproduce earlier results, we introduce another training paradigm which we term "prescient", where both $D_1$ and $D_2$ are known at all times, and which aligns well with evaluation methods in recent works.
As classifiers, we use typical DNN models like fully-connected- (fc), convolutional- (conv), LWTA-based- (fc-LWTA) and DNNs based on the EWC model (EWC).
Most of these can be combined with the concepts of Dropout (D, \cite{hinton2012improving}).
An overview of possible combinations is given in \cref{tab:combinations}.
\par
For all models, hyperparameter optimization is conducted in order to ensure that our results are not simply accidental.
\begin{table}[t!]
  \caption{
    Overview over 6 DNN models used in this study.
    They are obtained by combining the concept of Dropout (D) with the basic DNN models: fully-connected (fc), convolutional (conv), LWTA and EWC.
  }\label{tab:combinations}
  \centering
  \begin{tabular}{|l|C{1.2cm}|C{1.2cm}|C{3cm}|C{3cm}|}
    \hline
    \backslashbox{concept}{model} &  fc  &  conv  &      LWTA      &     EWC     \\ \hline
    with Dropout                  & D-fc & D-conv &   \ding{55}    & D-EWC (EWC) \\ \hline
    without Dropout               &  fc  &  conv  & LWTA-fc (LWTA) &  \ding{55}  \\ \hline
  \end{tabular}
\end{table}
\subsection{Related work on CF in DNNs}\label{sec:relwork}
In addition to early works on CF in connectionist models \cite{french}, new approaches specific to DNNs have recently been unveiled, some with the explicit goal of preventing catastrophic forgetting \cite{goodfellow2013empirical,kirkpatrick2017overcoming,srivastava2013compete,lee2017overcoming}. The work presented in \cite{goodfellow2013empirical} advocates the popular Dropout method as a means to reduce or eliminate CF, validating their claims on tasks derived a randomly shuffled version of MNIST \cite{lecun-01a} and a Sentiment Analysis problem.
In \cite{srivastava2013compete}, a new kind of competitive transfer function is presented which is termed LWTA (Local Winner Takes All).
In a very recent article \cite{kirkpatrick2017overcoming}, the authors advocate determining the hidden layer weights that are most "relevant" to a DNNs performance, and punishing the change of those weights more heavily during retraining by an additional term in the energy functional. Experiments are conducted on random data, randomly shuffled MNIST data as in \cite{goodfellow2013empirical,srivastava2013compete}, and on a task derived from Deep Q-learning in Atari Games \cite{mnih2015human}. 
Even more recently, authors in \cite{lee2017overcoming} propose the so-called incremental moment matching (IMM) technique which suggests an alignment of statistical properties of the DNN between $D_1$ and $D_2$ which is not included here, because it inherently requires knowledge of $D_1$ at re-training time to select the best regularization parameter(s).
\section{Methods}
\label{sec:methods}
The principal dataset this investigation is based on is MNIST \cite{lecun-01a}.
Despite being a very old benchmark, and a very simple one, it is still widely used, in particular in recent works on incremental learning in DNNs \cite{goodfellow2013empirical,kirkpatrick2017overcoming,lee2017overcoming,srivastava2013compete}.
It is used here because we wish to reproduce these results, and also because we care about performance in class-incremental settings, not offline performance on the whole dataset.
As we will see, MNIST-derived problems are more than a sufficient challenge for the tested algorithms, so it is really unnecessary to add more complex ones (but see \cref{sec:weak} for a more in-depth discussion of this issue).
\subsection {Learning tasks}\label{sec:methods:tasks}
As outlined in \cref{sec:intro:approach}, incremental learning performance of a given model is evaluated on several datasets constructed from the MNIST dataset.
The model is trained successively on two sub-tasks ($D_1$ and $D_2$) from the chosen dataset and it is recorded to what extend knowledge about previous sub-tasks is retained.
The precise way the sub-tasks of all datasets are constructed from the MNIST dataset shall be described below. \\
\textbf{Exclusion: D5-5} These datasets are obtained by randomly choosing 5 MNIST classes for $D_1$, and the remaining 5 for $D_2$.
To verify that results do not depend on a particular choice of classes, we create a total of 8 datasets where the partitioning of classes is different (see \cref{tab:datasets}). \\
\textbf{Exclusion: D9-1} We construct these datasets in a similar way as D5-5, selecting 9 MNIST classes for $D_1$ and the remaining class for $D_2$. In order to make sure that no artifacts are introduced, we create three datasets (D9-1a, D9-1b and D9-1c) with different choices for $D_1$ and $D_2$, see \cref{tab:datasets}. \\
\textbf{Permutation: DP10-10} This is the dataset used to evaluate incremental retraining in \cite{goodfellow2013empirical,kirkpatrick2017overcoming,srivastava2013compete}, so results can directly be compared.
It contains two sub-tasks, each of which is obtained by permuting each 28\,x\,28 image in a random fashion that is different between, but identical within, sub-tasks.
Since both sub-tasks contain 10 MNIST classes, we denote this dataset by DP10-10, the "P" indicating permutation, see \cref{tab:datasets}.
\setlength{\tabcolsep}{1.9mm}
\begin{table*}[b!]
  \caption{
    MNIST-derived datasets (DS) used in this article.
    All partitions of MNIST into $D_1$ and $D_2$ are non-overlapping.
    For the DP10-10 dataset, the classes are identical for $D_1$ and $D_2$ but pixels are permuted in $D_2$ as described in the text.
  }\label{tab:datasets}
  \centering
  \resizebox{\textwidth}{!}{%
    \begin{tabular}{|c|cccccccc|ccc|c|}
      \hline
      \multirow{2}{*}{\backslashbox{part.}{DS}} &                                                               \multicolumn{8}{c}{D5-5}                                                                & \multicolumn{3}{|c|}{D9-1} & \multirow{2}{*}{\footnotesize DP10-10} \\ \cline{2-12}
                                                & \multicolumn{1}{c}{D5-5a} &     D5-5b     &     D5-5c     &     D5-5d     &     D5-5e     &     D5-5f     &     D5-5g     & \multicolumn{1}{c|}{D5-h} & D9-1a & D9-1b &   D9-1c    &                                        \\ \hline
                    $D_1$ classes               &            0-4            & 0\,2\,4\,6\,8 & 3\,4\,6\,8\,9 & 0\,2\,5\,6\,7 & 0\,1\,3\,4\,5 & 0\,3\,4\,8\,9 & 0\,5\,6\,7\,8 &       0\,2\,3\,6\,8       &  0-8  &  1-9  &   0,2-9    &                  0-9                   \\
                    $D_2$ classes               &            5-9            & 1\,3\,5\,7\,9 & 0\,1\,2\,5\,7 & 1\,3\,4\,8\,9 & 2\,6\,7\,8\,9 & 1\,2\,5\,6\,7 & 1\,2\,3\,4\,9 &       1\,4\,5\,7\,9       &   9   &   0   &     1      &                  0-9                   \\ \hline
    \end{tabular}}
\end{table*}
\subsection{Models}\label{sec:methods:models}
We use TensorFlow/Python to implement or re-create all models used in this article.
The source code for all experiments is available at \url{https://gitlab.informatik.hs-fulda.de/ML-Projects/CF_in_DNNs}. \\ \\
\textbf{Fully connected deep network} Here, we consider a "normal" fully-connected (FC) feed-forward MLP with two hidden layers, a softmax (SM) readout layer trained using cross-entropy, and the (optional) application of Dropout (D) and ReLU operations after each hidden layer.
Its structure can thus be summarized as In-FC1-D-ReLU-FC2-D-ReLU-FC3-SM.
In case more hidden layers are added, their structure is analogous. \\
\textbf{ConvNet} A convolutional network inspired by \cite{ciresan2011flexible} is used here, with two hidden layers and the application of Dropout (D), max-pooling (MP) and ReLU after each layer, as well as a softmax (SM) readout layer trained using cross-entropy.
It structure can thus be stated as In-C1-MP-D-ReLU-C2-MP-D-ReLU-FC3-SM.\\
\textbf{EWC} The Elastic Weight Consolidation (EWC) model has been recently proposed in \cite{kirkpatrick2017overcoming} to address the issue of CF in incremental learning tasks.
We use a TensorFlow-implementation provided by the authors that we integrate into our own experimental setup; the corresponding code is available for download as described.
The basic network structure is analogous to that of fc models.\\
\textbf{LWTA} Deep learning with a fully-connected Locally-Winner-Takes-All (LWTA) transfer function has been proposed in \cite{srivastava2013compete}, where it is also suggested that deep LWTA networks have a significant robustness when trained incrementally with several tasks.
We use a self-coded TensorFlow implementation of the model proposed in \cite{srivastava2013compete}. Following \cite{srivastava2013compete}, the number of LWTA blocks is always set to 2. The basic network structure is analogous to that of fully-connected models. \\
\textbf{Dropout} Dropout, introduced in \cite{hinton2012improving} and widely used in recent research on DNNs, is a special transfer function that sets a random subset of activities in each layer to 0 during training.
It can, in principle, be applied to any DNN and thus can be combined with all previously listed models except EWC (already incorporated) and LWTA (unclear whether this would be sensible as LWTA is already a kind of transfer function).
\subsection{Experimental procedure}\label{sec:exp:proc}
The procedure we employ for all experiments is essentially the one given in \cref{sec:intro:approach}, where all models listed in \cref{sec:methods:models} and \cref{tab:combinations} are applied to a subset of class-incremental learning tasks described in \cref{sec:methods:tasks}.
For each experiment, characterized by a pair of model and task, we conduct a search in model parameter space for the best model configuration, leading to multiple runs per experiment, each run corresponding to a particular set of parameters for a given model and a given task.
\par
Each run lasts for $2t_{\textrm{max}}$ iterations and is structured as shown in \cref{fig:experiments}, initially training the chosen model first on sub-task $D_1$ and subsequently on sub-task $D_2$, each time for $t_{\textrm{max}}$ iterations.
Classification accuracy, measured at iteration $t$, on a test set $\mathcal{B}$ while training on a train set $\mathcal{A}$, is denoted $\chi(\mathcal{A},\mathcal{B},t)$.
For a thorough evaluation, we record the quantities $\chi(D_1, D_1, t<t_{\textrm{max}})$, $\chi(D_2,D_2,t\ge t_{\textrm{max}})$ and $\chi(D_2,D_1\cup D_2,t\ge t_{\textrm{max}})$.
Finally, the best-suited parameterized model must be chosen among all the runs of an experiment.
We investigate two strategies for doing this, corresponding to different levels of knowledge at training and retraining time during a single run.
As detailed in \cref{sec:intro:approach}, these are the strategies which we term "prescient" and "realistic".
The "prescient" evaluation strategy (see \cref{algo:prescient}) corresponds to an a priori knowledge of sub-task $D_2$ at initial training time, as well as to a knowledge about $D_1$ at retraining time.
Both assumptions are difficult to reconcile with incremental training in applied scenarios, as detailed in \cref{sec:intro:approach}.
We use this strategy here to compare our results to previous works in the field \cite{goodfellow2013empirical,kirkpatrick2017overcoming,srivastava2013compete}.
In contrast, the "realistic" evaluation strategy (see \cref{algo:realistic}) assumes no knowledge about future sub-tasks ($D_2$) and furthermore supposes that $D_1$ is unavailable at retraining time due to its size (see \cref{sec:intro:approach} for the reasoning).
It is this strategy which we propose for future investigations concerning incremental learning.
\subsection{Hyperparameters and model selection}\label{sec:exp:selection}
For runs from all experiments, not involving CNNs, the parameters that are varied are: number of hidden layers $L\in\{2,3\}$, layer sizes $S\in\{200,400,800\}$, learning rate during initial training $\epsilon_{D_1}\in\{0.01,0.001\}$, 
and learning rate du\-ring retraining $\epsilon_{D_2}\in\{0.001,0.0001,0.00001\}$.
Based on the parameter set $\mathcal{P} \subseteq L \times S \times \epsilon_{D_1} \times \epsilon_{D_2}$, all models are evaluated, respectively are model-specific hyper-parameters used or supplanted.
For experiments using CNNs, we fix the topology to a form known to achieve good performances on MNIST as an exhaustive optimization of all relevant parameters would prove too time-consuming in this case, and vary only the $\epsilon_{D_1}$ and $\epsilon_{D_2}$ as detailed before.
For EWC experiments, the importance parameter $\lambda$ of the retraining run is fixed at ${1}/{\epsilon_{D_2}}$, this choice is nowhere to be found in \cite{kirkpatrick2017overcoming} but is used in the provided code, which is why we adopt it.
For LWTA experiments, the number of LWTA blocks is fixed to $2$ in all experiments, corresponding to the values used in \cite{srivastava2013compete}.
Dropout rates, if applied, are set to $0.2$ (input layer) and $0.5$ (hidden layers), consistent with the choices made in \cite{goodfellow2013empirical}.
For CNNs, only a single Dropout rate of $0.5$ is applied for input and hidden layers alike.
The length $t_{\textrm{max}}$ of training\,/\,retraining period is empirically fixed to $2500$ iterations, each iteration using a batch size of $100$ ($\textrm{batch}_{\textrm{size}}$).
The Momentum optimizer provided by TensorFlow is used for performing training, with a momentum parameter $\mu=0.99$.
\subsection{Reproduction of previous results by prescient evaluation}
In this experiment, we wish to determine whether it is possible to find a parameterization for a given DNN model and task when there is a perfect knowledge about and availability of the initial and future sub-tasks.
Applying the models listed in \cref{sec:methods:models} to the tasks described in \cref{sec:methods:tasks}, and using the experimental procedure detailed in \cref{sec:exp:proc}, we obtain the results summarized in \cref{tab:dnn1} (applying the "prescient" evaluation of \cref{algo:prescient}).
\begin{algorithm}[b!]
\SetEndCharOfAlgoLine{} 
\SetAlgoVlined 
\SetAlgorithmName{Alg. \normalfont}{}{}
\SetAlgoCaptionSeparator{\normalfont{:}}
\KwData {model $m$, sub-tasks $D_1$\,\&\,$D_2$, parameter set $\mathcal{P}$}
\KwResult{quality of best model $q_{m_{\vec{p}}}^*$}
initialize $q_{m_{\vec{p}}}^* \gets -1$ \\
\ForEach{parameters $\vec{p} \in \mathcal{P}$} {
  initial training of $m_{\vec{p}}$ on $D_1$ for $t_{\textrm{max}}$ iterations \\
  \For(\hspace{2.2cm}{\textit{// retraining of $m_{\vec{p}}$ on $D_2$}}){$t \gets 0$ \KwTo $t_{\textrm{max}}$ iterations}{
    update $m_{\vec{p}}$ on $D_2$ using $\textrm{batch}_{\textrm{size}}$ \\ 
     $q_{m_{\vec{p}},t} \gets \chi(D_2,D_1\cup D_2,t)$ \\
    \lIf{$q_{m_{\vec{p}},t} > q_{m_{\vec{p}}}^*$}{
      $q_{m_{\vec{p}}}^* \gets q_{m_{\vec{p}},t}$
    }
  }
}
\KwRet $q_{m_{\vec{p}}}^*$
\caption{The \textit{prescient evaluation} strategy.}
\label{algo:prescient}
\end{algorithm}
We can state the following insights: first of all, we can reproduce the basic results from \cite{goodfellow2013empirical} using the fc model on DP10-10, which avoids catastrophic forgetting (contrarily to the conclusions drawn in this paper: these authors consider the very modest decrease in performance to be catastrophic forgetting).
This is however very specific to this particular task, and in fact all models except EWC exhibit blatant catastrophic forgetting behavior particularly on the D5-5 type tasks, while performing ad\-equate\-ly if not perfectly on the D9-1 tasks.
EWC performs well on these tasks as well, so we can state that EWC is the only tested algorithm that avoids CF for all tasks when using prescient evaluation.
Another observation is that the use of Dropout, as suggested in \cite{goodfellow2013empirical}, does not seem to significantly improve matters.
The LWTA method performs a little better than fc, D-fc, conv and D-conv but is surpassed by EWC by a very large margin.
\setlength{\tabcolsep}{2.61mm}
\begin{table*}[t!]
    \caption{
      Results for \textbf{prescient evaluation}.
      Please note that the performance level of complete catastrophic forgetting (i.e., chance-level classification after retraining with $D_2$) depends on the dataset considered: for the D5-5 dataset it is at 0.5, whereas it is at 0.1 for the D9-1 datasets.
      The rightmost column indicates the DP10-10 task which is solved near-perfectly by all models.
    }\label{tab:dnn1}
  \centering
  \resizebox{\textwidth}{!}{%
  \begin{tabular}{|c||cccccccc|ccc|c|}
    \hline
    \multirow{2}{*}{\backslashbox{model}{dataset}} &                   \multicolumn{8}{c}{D5-5}                    & \multicolumn{3}{|c|}{D9-1} & \multirow{2}{*}{DP10-10} \\ \cline{2-12}
                                                   & D5-5a & D5-5b & D5-5c & D5-5d & D5-5e & D5-5f & D5-5g & D5-5h & D9-1a & D9-1b &   D9-1c    &                          \\ \hline
                         EWC                       & 0.92  & 0.92  & 0.91  & 0.93  & 0.94  & 0.94  & 0.89  & 0.93  & 1.00  & 1.00  &    1.00    &           1.00           \\
                          fc                       & 0.69  & 0.63  & 0.58  & 0.65  & 0.61  & 0.58  & 0.61  & 0.69  & 0.87  & 0.87  &    0.86    &           0.97           \\
                         D-fc                      & 0.58  & 0.60  & 0.61  & 0.66  & 0.61  & 0.54  & 0.63  & 0.64  & 0.87  & 0.87  &    0.85    &           0.96           \\
                         conv                      & 0.51  & 0.50  & 0.50  & 0.50  & 0.50  & 0.50  & 0.51  & 0.49  & 0.89  & 0.89  &    0.87    &           0.95           \\
                        D-conv                     & 0.51  & 0.50  & 0.50  & 0.50  & 0.50  & 0.50  & 0.50  & 0.49  & 0.81  & 0.84  &    0.87    &           0.96           \\
                         LWTA                      & 0.66  & 0.68  & 0.64  & 0.73  & 0.71  & 0.62  & 0.68  & 0.71  & 0.88  & 0.91  &    0.91    &           0.97           \\ \hline
  \end{tabular}}
\end{table*}
\subsection{Realistic evaluation}
\begin{figure*}[h!]
  \centering
  \vspace{-0.5cm}
  \subfloat[Task: D9-1a]{\includegraphics[width=0.45\textwidth, trim=0cm 2.5cm 0cm 4.0cm, clip]{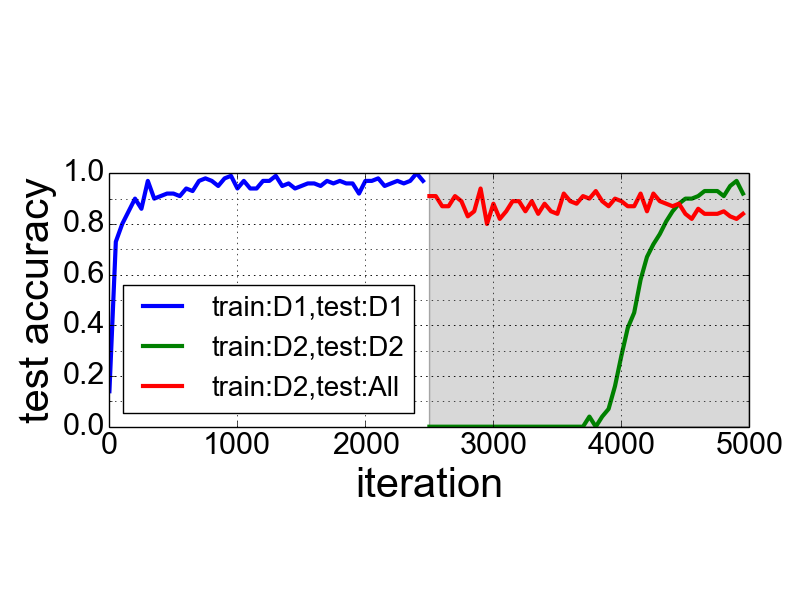}}
  \subfloat[Task: D9-1c]{\includegraphics[width=0.45\textwidth, trim=0cm 2.5cm 0cm 4.0cm, clip]{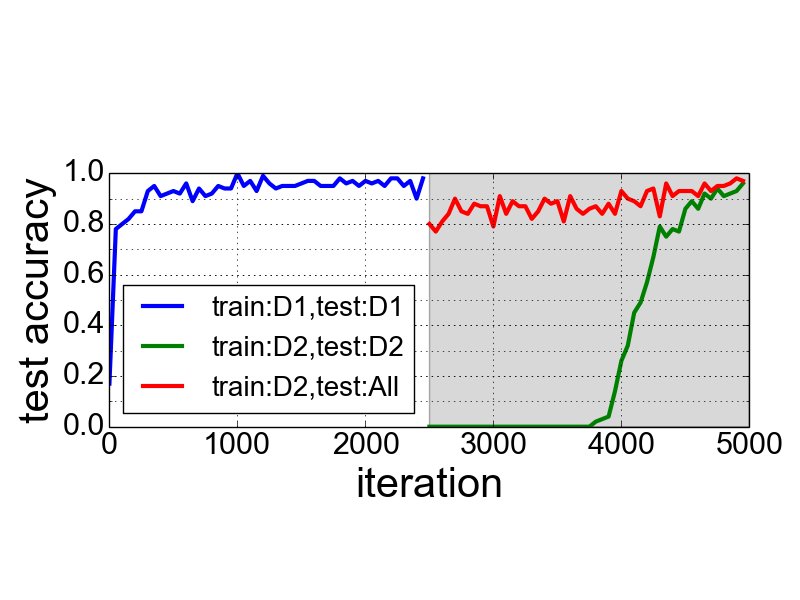}}
  \caption{Best EWC runs on D9-1 datasets in the \textbf{realistic evaluation} condition.}\label{fig:plotsEWC1}
  \centering
  \subfloat[Task: D5-5a]{\includegraphics[width=0.45\textwidth, trim=0cm 2.5cm 0cm 4.0cm, clip]{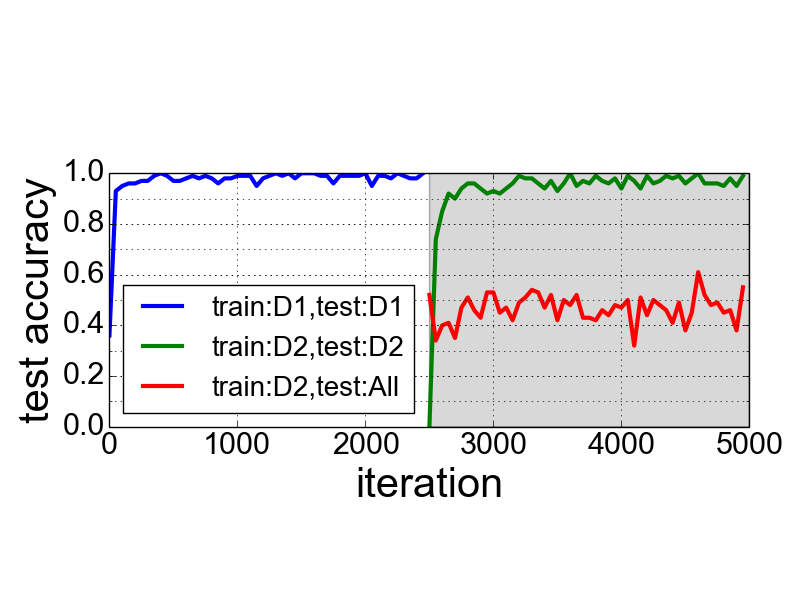}}
  \subfloat[Task: D5-5h]{\includegraphics[width=0.45\textwidth, trim=0cm 2.5cm 0cm 4.0cm, clip]{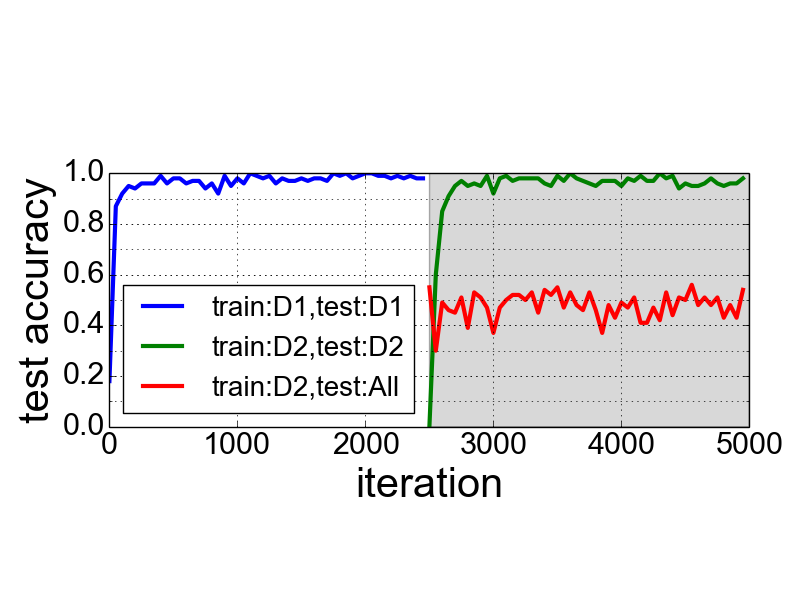}}
  \caption{Best EWC runs on D5-5 datasets in the \textbf{realistic evaluation} condition.}\label{fig:plotsEWC2}
  \centering
  \subfloat[Task: D9-1b]{\includegraphics[width=0.45\textwidth, trim=0cm 2.5cm 0cm 4.0cm, clip]{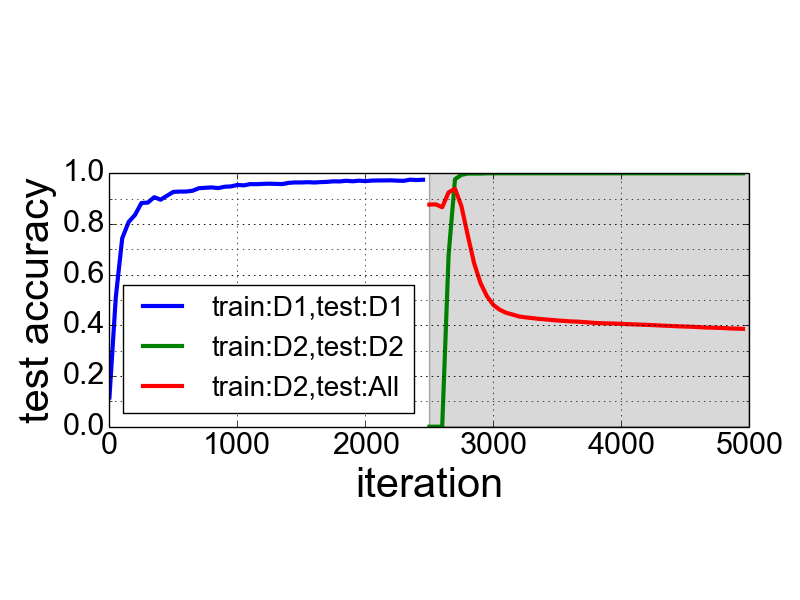}}
  \subfloat[Task: D9-1c]{\includegraphics[width=0.45\textwidth, trim=0cm 2.5cm 0cm 4.0cm, clip]{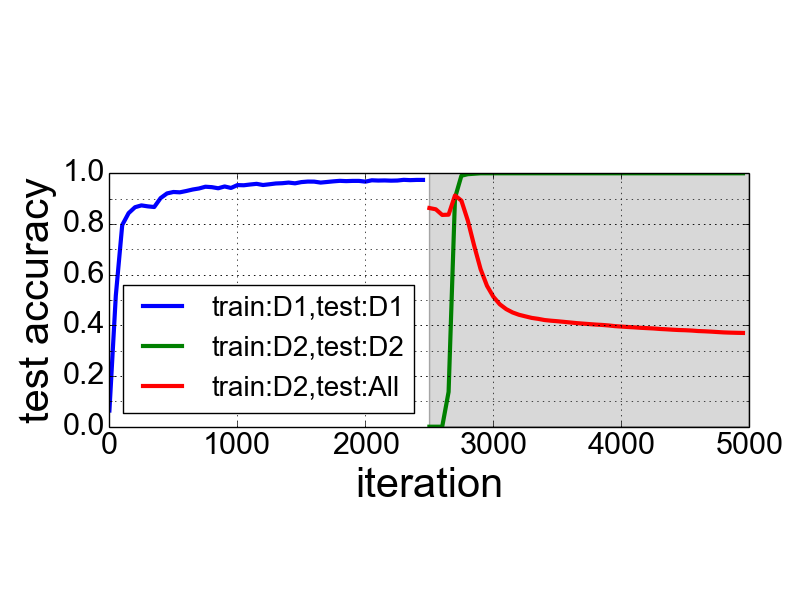}}
  \caption{Best D-fc runs on D9-1 datasets in the \textbf{realistic evaluation} condition.}\label{fig:plotsDfc1}
  \centering
  \subfloat[Task: D5-5a]{\includegraphics[width=0.45\textwidth, trim=0cm 2.5cm 0cm 4.0cm, clip]{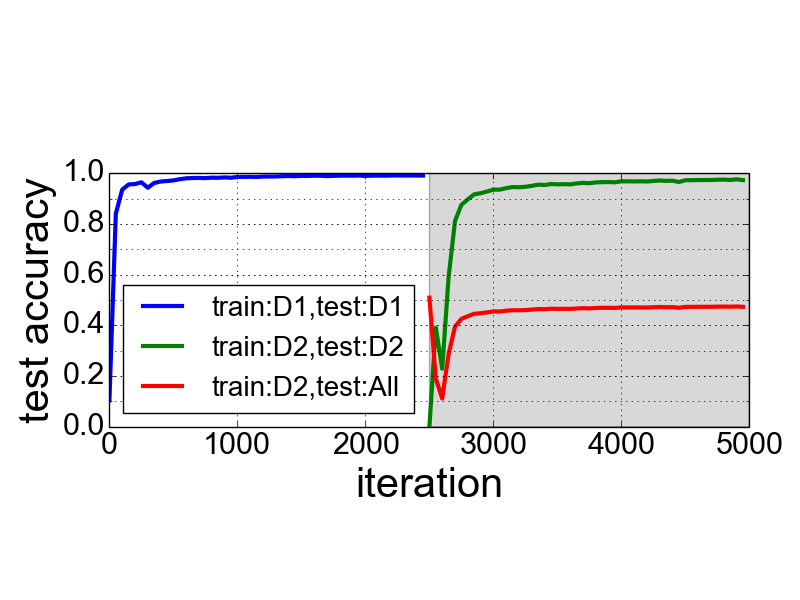}}
  \subfloat[Task: D5-5c]{\includegraphics[width=0.45\textwidth, trim=0cm 2.5cm 0cm 4.0cm, clip]{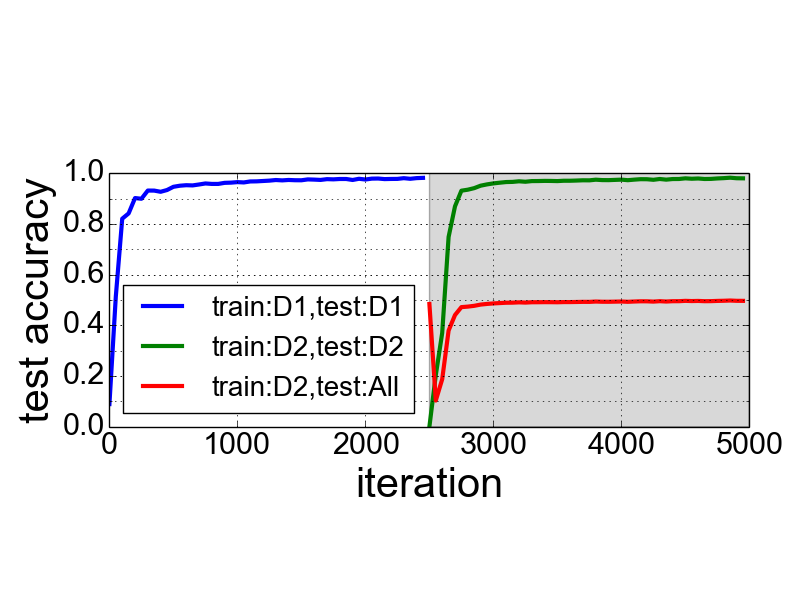}}
  \caption{Best D-fc runs on D5-5 datasets in the \textbf{realistic evaluation} condition.}\label{fig:plotsDfc2}
\end{figure*}

\begin{table*}[b!]
  \caption{
    Results for \textbf{realistic evaluation}.
    Please note that the performance level of total catastrophic forgetting (i.e., chance-level classification after retraining with $D_2$) depends on the dataset: for the D5-5 dataset it is at 0.5, whereas it is at 0.1 for the \mbox{D9-1} datasets.
    The rightmost column indicates the DP10-10 task ("permuted MNIST") which is again solved near-perfectly by all models.
  }\label{tab:dnn3}
  \centering
  \resizebox{\textwidth}{!}{%
    \begin{tabular}{|c||cccccccc|ccc|c|}
      \hline
      \multirow{2}{*}{\backslashbox{model}{dataset}} &                   \multicolumn{8}{c}{D5-5}                    & \multicolumn{3}{|c|}{D9-1} & \multirow{2}{*}{DP10-10} \\ \cline{2-12}
      & D5-5a & D5-5b & D5-5c & D5-5d & D5-5e & D5-5f & D5-5g & D5-5h & D9-1a & D9-1b &   D9-1c    &                          \\ \hline
      EWC                       & 0.48  & 0.56  & 0.62  & 0.52  & 0.58  & 0.58  & 0.55  & 0.53  & 0.82  & 0.91  &    0.97    &           0.99           \\
      fc                       & 0.47  & 0.49  & 0.50  & 0.50  & 0.48  & 0.49  & 0.50  & 0.49  & 0.15  & 0.10  &    0.23    &           0.97           \\
      D-fc                      & 0.47  & 0.50  & 0.50  & 0.50  & 0.49  & 0.49  & 0.50  & 0.49  & 0.52  & 0.10  &    0.16    &           0.96           \\
      conv                      & 0.48  & 0.50  & 0.50  & 0.50  & 0.49  & 0.50  & 0.51  & 0.49  & 0.29  & 0.33  &    0.11    &           0.95           \\
      D-conv                     & 0.48  & 0.50  & 0.50  & 0.50  & 0.45  & 0.50  & 0.50  & 0.49  & 0.24  & 0.22  &    0.14    &           0.96           \\
      LWTA                      & 0.47  & 0.50  & 0.50  & 0.50  & 0.49  & 0.49  & 0.51  & 0.49  & 0.48  & 0.29  &    0.66    &           0.97           \\ \hline
  \end{tabular}}
\end{table*}
This experiment imposes the much more restrictive/realistic evaluation, detailed in \cref{sec:exp:proc} and \cref{algo:realistic}, essentially performing initial training and model selection only on $D_1$ and retraining only using $D_2$.
It is this or related schemes that would have to be used in typical application scenarios, and thus represents the principal subject of this article.
The performances of all tested DNN models on all of the tasks from \cref{sec:methods:tasks} are summarized in \cref{tab:dnn3}.
Plots of experimental results over time for the D-fc and EWC models are given in \crefrange{fig:plotsEWC1}{fig:plotsDfc2}.
The results show a rather bleak picture where only the EWC model achieves significant success for the D9-1 type tasks while failing for the D5-5 tasks.
All other models do not even achieve this partial success and exhibit strong CF for all tasks.
We can therefore observe that a different choice of evaluation procedure strongly impacts results and the conclusions which are drawn concerning CF in DNNs.
For the realistic evaluation condition, which in our view is much more relevant than the prescient one used in nearly all of the related work on the subject, CF occurs for all DNN models we tested, and partly even for the EWC model.
As to the question why EWC performs well for all of the D9-1 type task in contrast to the D5-5 type tasks, one might speculate that the addition of five new classes, as opposed to one, might exceed EWC's capabilities of protecting the weights most relevant to $D_1$.
Various different values of the constant $\lambda$ governing the contribution of Fisher information in EWC were tested but with very similar results.
\begin{algorithm}[t!]
\SetEndCharOfAlgoLine{}
\SetAlgoVlined 
\SetAlgorithmName{Alg.\normalfont}{}{}
\SetAlgoCaptionSeparator{\normalfont{:}}
\KwData {model $m$, sub-tasks $D_1$\,\&\,$D_2$, parameter set $\mathcal{P}$}
\KwResult{quality of best model $q_{m_{\vec{p}}}^*$}
initialize $q_T^* \leftarrow -1$ \\
\ForAll(\hspace{0.6cm}\textit{//determine best model parameter training $D_1$}){parameters $\vec{p} \in \mathcal{P}$} {\texttt{}
  \For{$t \gets 0$ \KwTo $t_{\textrm{max}}$ iterations}{
    update of $m_{\vec{p}}$ on $D_1$ using $\textrm{batch}_{\textrm{size}}$;  $q_{m_{\vec{p}},t} \gets \chi(D_1, D_1 ,t)$ \\
  \lIf{$q_{m_{\vec{p}},t} > q_T^* $}{
    $q_T^* \gets q_{m_{\vec{p}},t}$;  $m_{\vec{p}}^* \gets m_{\vec{p}}$
  }
  }
}
initialize $q_{m_{\vec{p}}}^* \gets -1$ \\
\ForAll{retraining learning rates $\epsilon \in \epsilon_{D_2}$} {
  initialize $q_R^* \leftarrow -1$ \\
  \For(\hspace{2.2cm}\textit{//retraining of $m_{\vec{p}}^*$ on $D_2$}){ $t \gets 0$ \KwTo $t_{\textrm{max}}$ iterations}{
    update $m_{\vec{p}}^*$ on $D_2$ with learning rate $\epsilon$; $q_{m_{\vec{p}},t} \gets \chi(D_2, D_2 ,t)$ \\
  \lIf{$q_{m_{\vec{p}},t} > q_R^*$}{
    $q_R^* \gets q_{m_{\vec{p}},t}$
    }
  }
  $t_E \gets \arg\min_t (q_{m_{\vec{p}},t} > 0.99 \cdot q_R^*)$;  $q_{m_{\vec{p}}} \gets \chi(D_2,D_1\cup D_2,t_E)$\\
  \lIf {$q_{m_{\vec{p}}} > q_{m_{\vec{p}}}^*$}{$q_{m_{\vec{p}}}^* \gets q_{m_{\vec{p}}}$}{}
}
\KwRet $q_{m_{\vec{p}}}^*$
\caption{The \textit{realistic evaluation} strategy.}
\label{algo:realistic}
\end{algorithm}
\captionsetup[subfigure]{labelformat=empty}
\section{Discussion of results and principal conclusions}
From our experiments, we draw the following principal conclusions:
\begin{itemize}[leftmargin=*,topsep=0.2cm]
  \setlength{\itemsep}{-0.0cm}
  \item CF should be investigated using the appropriate evaluation paradigms that reflect application conditions.
  At the very least, using future data for model selection is inappropriate, which leads to conclusions that are radically different from most related experimental work, see \cref{sec:relwork}.
  \item using a realistic evaluation paradigm, we find that CF is still very much a problem for all investigated methods.
  \item in particular: Dropout is not effective against CF; neither is LWTA.
  \item the permuted MNIST task can be solved by almost any DNN model in almost any topology.
  So all conclusions drawn from using this task should be revisited.
  \item EWC seems to be partly effective but fails for all of the D5-5 tasks, indicating that it is not the last word in this matter.
\end{itemize}
\vspace{1em}
We write that EWC "seems to be partly effective", meaning it solves some incremental tasks well while it fails for others.
So we observe that there is no guarantee that can be obtained from a purely empirical validation approach such as ours; yet another type of incremental learning task might be solved perfectly or not at all.
This points to the principal conceptual problem that we see when investigating CF in DNNs: there is no theory that might offer any guarantees.
Such guarantees could be very useful in practice, the most interesting one being how to determine a lower bound on performance loss on $D_1$ $\cup$ $D_2$, without having access to $D_1$, only to the network state and $D_2$.
Other guarantees could provide upper bounds on retraining time before performance on $D_1$ $\cup$ $D_2$ degrades.
\section{Future work}\label{sec:weak}
The issue of CF is a complex one, and correspondingly our article and our experimental procedures are complex as well.
There are several points where we made rather arbitrary choices, e.g., when choosing the constant $\mu=0.99$ in the realistic evaluation \cref{algo:realistic}.
The results are affected by this choice although we verified that the trend is unchanged.
Another weak point is our model selection procedure: a much larger combinatorial set of model hyper-parameters should be sampled, including Dropout rates, convolution filter kernels, number and size of layers.
This might conceivably allow to identify model hyperparameters avoiding CF for some or all tested models, although we consider this unlikely.
Lastly, the use of MNIST might be criticized as being too simple: this is correct, and we are currently doing experiments with more complex classification tasks (e.g., SVHN and CIFAR-10). However, as our conclusion is that none of the currently proposed DNN models can avoid CF, this is not very likely to change when using an even more challenging classification task (rather the reverse, in fact).
%
{\small
\bibliographystyle{splncs04}
\bibliography{icann18}
}
The final authenticated version is available online at \url{https://doi.org/10.1007/978-3-030-01418-6_48}.
\end{document}

%% file: figs/figExperiment2.pdf_tex
\begingroup%
  \makeatletter%
  \providecommand\color[2][]{%
    \errmessage{(Inkscape) Color is used for the text in Inkscape, but the package 'color.sty' is not loaded}%
    \renewcommand\color[2][]{}%
  }%
  \providecommand\transparent[1]{%
    \errmessage{(Inkscape) Transparency is used (non-zero) for the text in Inkscape, but the package 'transparent.sty' is not loaded}%
    \renewcommand\transparent[1]{}%
  }%
  \providecommand\rotatebox[2]{#2}%
  \newcommand*\fsize{\dimexpr\f@size pt\relax}%
  \newcommand*\lineheight[1]{\fontsize{\fsize}{#1\fsize}\selectfont}%
  \ifx\svgwidth\undefined%
    \setlength{\unitlength}{272.70787036bp}%
    \ifx\svgscale\undefined%
      \relax%
    \else%
      \setlength{\unitlength}{\unitlength * \real{\svgscale}}%
    \fi%
  \else%
    \setlength{\unitlength}{\svgwidth}%
  \fi%
  \global\let\svgwidth\undefined%
  \global\let\svgscale\undefined%
  \makeatother%
  \begin{picture}(1,0.48469764)%
    \lineheight{1}%
    \setlength\tabcolsep{0pt}%
    \put(0.69052652,0.45298981){\color[rgb]{0,0,0}\makebox(0,0)[t]{\lineheight{0}\smash{\begin{tabular}[t]{c}\textbf{sub-task $\mathbf{D_2}$}\end{tabular}}}}%
    \put(-0.00277527,0.07081109){\color[rgb]{0,0,0}\makebox(0,0)[lt]{\lineheight{0}\smash{\begin{tabular}[t]{l}$0$\end{tabular}}}}%
    \put(0,0){\includegraphics[width=\unitlength,page=1]{figs/figExperiment2.pdf}}%
    \put(0.46019259,0.07081109){\color[rgb]{0,0,0}\makebox(0,0)[t]{\lineheight{0}\smash{\begin{tabular}[t]{c}$t_{max}$\end{tabular}}}}%
    \put(0.81903123,0.07081109){\color[rgb]{0,0,0}\makebox(0,0)[lt]{\lineheight{0}\smash{\begin{tabular}[t]{l}$2t_{max}$\end{tabular}}}}%
    \put(0.45728304,0.00053285){\color[rgb]{0,0,0}\makebox(0,0)[t]{\lineheight{0}\smash{\begin{tabular}[t]{c}iterations\end{tabular}}}}%
    \put(0,0){\includegraphics[width=\unitlength,page=2]{figs/figExperiment2.pdf}}%
    \put(0.23337522,0.3840813){\color[rgb]{0,0,0}\makebox(0,0)[t]{\lineheight{0}\smash{\begin{tabular}[t]{c}train on $\mathbf{D_1}$\end{tabular}}}}%
    \put(0.23244739,0.29974196){\color[rgb]{0,0,0}\makebox(0,0)[t]{\lineheight{0}\smash{\begin{tabular}[t]{c}test on $\mathbf{D_1}$\end{tabular}}}}%
    \put(0.69181567,0.3840813){\color[rgb]{0,0,0}\makebox(0,0)[t]{\lineheight{0}\smash{\begin{tabular}[t]{c}train on $\mathbf{D_2}$\end{tabular}}}}%
    \put(0.69108219,0.29974196){\color[rgb]{0,0,0}\makebox(0,0)[t]{\lineheight{0}\smash{\begin{tabular}[t]{c}test on $\mathbf{D_2}$\end{tabular}}}}%
    \put(0.69155784,0.21246909){\color[rgb]{0,0,0}\makebox(0,0)[t]{\lineheight{0}\smash{\begin{tabular}[t]{c}test on $\mathbf{D_1}$\end{tabular}}}}%
    \put(0.23142537,0.45298981){\color[rgb]{0,0,0}\makebox(0,0)[t]{\lineheight{0}\smash{\begin{tabular}[t]{c}\textbf{sub-task $\mathbf{D_1}$}\end{tabular}}}}%
    \put(0,0){\includegraphics[width=\unitlength,page=3]{figs/figExperiment2.pdf}}%
  \end{picture}%
\endgroup%